\documentclass[runningheads]{llncs}
\usepackage[T1]{fontenc}
\usepackage{graphicx}
\usepackage{float}
\usepackage[binary-units=true]{siunitx}
\usepackage[colorlinks,bookmarksopen,bookmarksnumbered,citecolor=red,urlcolor=red]{hyperref}
\usepackage{amsmath}
\usepackage{amsfonts}
\usepackage{amssymb}
\usepackage{footmisc}
\usepackage[capitalise]{cleveref}
\usepackage{siunitx}
\usepackage{array}
\usepackage{todonotes}
\usepackage{tikz}
\usetikzlibrary{arrows,shapes,positioning,fit} 
\usepackage[numbers,sectionbib]{natbib}
\usepackage{subcaption}
\usepackage{booktabs}

\tikzset{%
  block/.style    = {draw, thick, rectangle, minimum height = 3em,
    minimum width = 3em}
}

\newcolumntype{L}[1]{>{\raggedright\arraybackslash}m{#1}} 
\newcolumntype{C}[1]{>{\centering\arraybackslash}m{#1}}   
\newcolumntype{R}[1]{>{\raggedleft\arraybackslash}m{#1}}  

\newcommand\blfootnote[1]{%
  \begingroup
  \renewcommand\thefootnote{}\footnote{#1}%
  \addtocounter{footnote}{-1}%
  \endgroup
}%

\begin{document}
\title{Comparing Computing Platforms for Deep Learning on a Humanoid Robot}
%
%
\author{Alexander Biddulph\thanks{Supported by an Australian Government Research Training Program scholarship and a
	top-up scholarship through 4Tel Pty.} \and
	Trent Houliston \and
	Alexandre Mendes \and
Stephan K. Chalup}
\authorrunning{A. Biddulph et al.}
%
\institute{School of Electrical Engineering and Computing \\
	The University of Newcastle, Callaghan, NSW, 2308, Australia. \\
	\email{Alexander.Biddulph@uon.edu.au}}

\maketitle 
\begin{abstract}
	The goal of this study is to test two different computing platforms with respect to their suitability for running
	deep networks as part of a humanoid robot software system. One of the platforms is the CPU-centered
	Intel\textregistered~NUC7i7BNH and the other is a NVIDIA\textregistered~Jetson TX2 system that puts more emphasis on
	GPU processing. The experiments addressed a number of benchmarking tasks including pedestrian detection using deep
	neural networks. Some of the results were unexpected but demonstrate that platforms exhibit both advantages and
	disadvantages when taking computational performance and electrical power requirements of such a system into account.

\end{abstract}

\section{Introduction}
\label{Introduction}
\blfootnote{The final authenticated publication is available online at \url{https://doi.org/10.1007/978-3-030-04239-4_11}}

Deep learning comes with challenges with respect to computational resources and training data requirements
\citep{GoodfellowEtAl2016,JabbarEtAl2017}. Some of the breakthroughs in deep neural networks (DNNs) only became possible
through the availability of massive computing systems or through careful co-design of software and hardware. For
example, the AlexNet system presented in~\citep{KrizhevskyEtAl2012} was implemented efficiently utilising two
NVIDIA\textregistered~GTX580 GPUs for training.

Machine learning on robots has been a growing area over the past
years~\citep{ChalupEtAlSMC2007,LevineEtAl2017,MetcalfeEtAl2016,PiersonGashler2017}. It has become increasingly desirable
to employ DNNs in low powered devices, among them humanoid robot systems, specifically for complex tasks such as object
detection, walk learning, and behaviour learning. Robot software systems that involve DNNs face the challenge of fitting
their large computational demands on a suitable computing platform that also complies with the robot's electrical power
budget. One way to address these challenges is to develop and use modern software architectures and efficient algorithms
for the most resource hungry components, e.g.\ computer vision~\citep{HoulistonEtAl2015}.

The need for hardware that can run deep convolutional neural networks (DCNNs) on low powered devices can be seen in
humanoid robotic systems that only possess CPU resources. Spec et al.~\cite{SpecEtAl2016} developed a DCNN designed to locate
soccer balls on a field. The final performance of this system was insufficient for practical use, with an execution time
of \SI{0.91}{\second} per frame. In a dynamic environment, both the framerate and latency of this detection would reduce
its usefulness. Furthermore, the developed architecture consumed too much RAM ($>$ \SI{2}{\giga\byte}) to be usable on
the chosen humanoid robot system.

Javadi et al.~\cite{JavadiEtAl2017} compared the performance of several well-known DCNN architectures on the task of detecting and
classifying other humanoid robots. The performance of these networks was compared both on GPUs and the CPUs of the
target robots. Even for the classification task, none of the tested deep network architectures achieved acceptable
performance. Only a two-layer network executed in less than \SI{100}{\milli\second} on their target platform.

While the design of the robot's brain and software system is the most crucial part of a humanoid robot system, every
other component of the robot, hardware or software, has to be carefully considered, as well. This paper compares a few
currently popular devices that can be used as the main computing device on a low-powered humanoid robot, with the aim to
allow usage of deep learning as part of the robot software system.

\section{Hardware Platforms}
In this section we will describe the robotic platform and two computers that can be used in similar autonomous systems.

The NUgus is a humanoid robotic platform designed to perform human-like activities in real-world environments. It stands
\SI{90}{\centi\meter} tall, weighs \SI{7.5}{\kilogram}, and can serve as an autonomous humanoid robot soccer player.
The original robotic platform was developed by the University of Bonn as an open platform in collaboration with the
company igus\textregistered~GmbH in Germany in 2015~\citep{AllgeuerEtAl2015}. The technical specifications of the NUgus
differ only slightly from the original design with the PC being replaced by an
Intel\textregistered~NUC7i7BNH~\citep{IntelNUC}, referred to as ``NUC'' in this paper, and the camera replaced with two
FLIR\textregistered~Flea\textregistered3 USB3 cameras fitted with \ang{195} wide angle lenses. The NUC features an
integrated GPU, which makes the platform suitable for complex computations, such as deep learning.




The NUgus's $20$ servos are the most power hungry component of the robot. Consuming up to \SI{225}{\watt}, the power
consumption of the computing platform makes up a relatively small portion of this power consumption. Despite this,
decreasing power consumption by utilising more efficient algorithms is an effective way of increasing battery life.

The second computer tested in this work is a NVIDIA\textregistered~Jetson TX2 system, referred to as ``Jetson'' in this
paper, is described in \citep{Franklin2017}. This system is now becoming a popular alternative for embedded systems that
require higher computing capacity, among them autonomous vehicles.



\section{The NUgus Robot Software System}
The NUgus control software system allows it to play soccer autonomously. It comprises several components; namely vision,
behaviour, kinematics and localisation. By far, the most complex component is the vision system, and it is also the most
computationally demanding. The four components run in parallel, and information must be shared among them in real time.
To achieve this goal the NUClear software framework is used~\citep{HoulistonEtAl2016}. When running DNNs in the robot
vision system, it is important that it does not over-utilise the available computing resources, or the robot will not be
able to perform the other activities necessary for playing soccer, e.g.\ walking while keeping itself balanced,
approaching the ball from the right direction and kicking the ball towards the goal.

\section{Deep Learning for Robot Vision}
The main aim of this work is to trial a DNN in the vision pipeline of a humanoid robotic system to determine the
feasibility of using DNNs in battery powered environments. To this end, SSD MobileNet~\citep{MobileNet,SSD} pretrained
on MS COCO~\citep{MSCOCO}, was acquired from the Tensorflow Model Zoo~\citep{TensorflowModelZoo}. This network is
reported to have a mAP of 21. Testing the network required its integration into the NUgus's vision system and the
implementation of an interface to allow communication through the NUbots software architecture.

\section{Experiments}
\label{sec:experiments}
In general, while the performance gains of GPU computing over CPU computing are undeniable the performance gain is not
always so significant, with GPU computing sometimes providing as little as a 2.5X increase over CPU
performance~\citep{lee2010debunking}. Gregg et al.~\cite{GreggEtAl2011} highlight the importance of considering data transfer times
to and from the GPU, as these transfer times can often dwarf the time spent performing the actual calculations making
GPU usage ineffective~\cite{AbouelfaragEtAl2017}.

Lee et al.~\cite{lee2010debunking} and Vanhoucke et al.~\cite{VanhouckeEtAl2011} emphasise that careful implementation and tuning using SIMD
intrinsics improves the performance of CPU-based algorithms, especially those using matrix multiplications.

This section details the experiments that were undertaken during the course of this work and is divided into two parts.
\cref{sec:exp_benchmarks} focuses on four CPUs and GPUs. Two of the CPUs and GPUs come as part of the NUC and Jetson
platforms, while the other two CPUs and GPUs are from a desktop and a laptop. In this section, the task was based on a
2D matrix rotation. Finally, in \cref{sec:exp_ped}, we present the results for the more complex task of pedestrian
detection. This task requires two image pre-processing steps and then the use of a DNN for the detection as a third
step. The pedestrian detection benchmark was carried out on the NUC and Jetson platforms only.

It should be noted that the tests presented in this section are not designed to provide a definitive performance
indication, but should be seen as a point of reference.

\subsection{CPU and GPU Benchmarking}
\label{sec:exp_benchmarks}
In order to determine a basis for comparing the NUC and the Jetson a simple benchmarking experiment was performed on the
CPU and GPU components of both devices. This experiment was also repeated on two consumer grade GPUs -- a low-end 
Intel\textregistered~HD Graphics 630 and a high-end NVIDIA\textregistered~GTX1080Ti -- to provide an idea of how the 
devices compare to end-user GPUs, and to ensure differences between OpenCL and CUDA implementations are evaluated on a
single device. Two consumer grade CPUs were also used -- an Intel~\textregistered~Core i7-7800X and i7-7920HQ -- to 
provide a similar comparison. \cref{tab:gpu_devices} lists the GPU devices and \cref{tab:cpu_devices} lists the CPU
devices that were used in this experiment.

The benchmarking experiment follows the experiment laid out in~\citep{GPUBenchkmarch}. The experiment is designed to
test the computational power, in terms of floating point operations per second (FLOPS), of a GPU without main memory
access.

The FLOPS measure is calculated using a 2D matrix multiplication operation. First, a 2D rotation matrix is created,
and a 2D vector is rotated multiple times. \cref{eqn:benchmark} shows the iterative formulation used for benchmarking
and was implemented in both OpenCL and CUDA for execution on the NUC and Jetson GPUs, respectively.

\begin{equation}
	\label{eqn:benchmark}
	\begin{split}
		\vec{R} &=
		\begin{bmatrix}
			\cos\left(2\right) & -\sin\left(2\right)   \\
			\sin\left(2\right) & ~~~\cos\left(2\right) \\
		\end{bmatrix}
	\end{split}
	\qquad\qquad
	\begin{split}
		\vec{x}_{0} &= \begin{bmatrix}1 & 0\end{bmatrix} \\
		\vec{x}_{1}^{T} &= \vec{R}\vec{x}_{0}^{T}
	\end{split}
	\qquad\qquad
	\begin{split}
		\vec{x}_{n}^{T} &= \vec{R}\vec{x}_{n - 1}^{T} \\
		out_{n} &= \langle\vec{x}_{n,0}, \vec{x}_{n,1}\rangle
	\end{split}
\end{equation}

The kernel used in this experiment has allowances for operating with varying dimensionality; mainly $1$, $2$, and
$4$-dimensional data. If $1$-dimensional data is rotating a single 2D point, then $2$-dimensional data would
amount to rotating two 2D points, and $4$-dimensional data is rotating four 2D points. Only $x_{0}$ in
\cref{eqn:benchmark} needs to be modified to account for changing dimensionality. The kernel also allows changing the
number of iterations that can be performed -- in our tests we set the number of iterations to $40,000$. The 
kernels were run with an increasing load on the GPU by changing the number of workers that are performing these
calculations. The number of workers ranged from $256$ to $1,024,000$ in steps of $256$.

Analysing \cref{eqn:benchmark} in more depth, each iteration contains $6$ floating point operations -- 
$4$ multiplications and $2$ additions. Factoring in the dimensionality of the data ($D$), the number of iterations
($N$), the number of workers ($W$), and the time taken to complete all iterations ($t$), we can derive a formula
for the number of floating point operations that the GPU can perform per second $FLOPS = \frac{6DNW}{t}$.

To ensure that the compiler does not try to optimise out any of the calculations, the final equation in
\cref{eqn:benchmark} is added at the end of the kernel. Strictly speaking, this modifies the formula for $FLOPS$ to
$\frac{6DNW}{t} + \frac{\left(2D - 1\right)W}{t}$. However, this extra term is insignificant and we only use the first
term in this work.

For the CPU benchmarking \cref{eqn:benchmark} was implemented using SIMD intrinsics. A varying number of tasks, from $1$
up $4,096$ in steps of $32$, were created using OpenMP.\@ All other parameters remained the same as in the GPU
benchmarking.

\subsection{Image Reprojection, Demosaicing, and Pedestrian Detection}
\label{sec:exp_ped}
\begin{figure}[htbp]
	\centering
	\scalebox{0.85}{
        \begin{tikzpicture}[every text node part/.style={align=center}]
            \node[block] (B) {DEMOSAICING \\ C++/OpenCL/CUDA \\ GPU};
            \node[block, right=2cm of B] (C) {REPROJECTION \\ C++/OpenCL/CUDA \\ GPU};
            \node[draw=black, fit=(B) (C)] (E) {};
            \node[block, above=of E] (A) {CAMERA \\ C++ \\ CPU};
            \node[block, below=of E] (D) {PEDESTRIAN DETECTOR \\ Python \\ GPU*};

            \draw[thick, ->] (A.south) -- (E.north) node[midway, right] {1280x1024 \\ Bayer BGGR};
            \draw[thick, <->] (B.east) -- (C.west) node[midway, above] {Concurrent};
            \draw[thick, ->] (E.south) -- (D.north) node[midway, right] {1280x1024 \\ RGB};
        \end{tikzpicture}
    }
	\caption{\label{fig:pipeline}
             The software pipeline of the pedestrian detection system. Images from the camera are streamed
             to the image demosaicing and reprojection module. After demosaicing and reprojection it is passed to the
             pedestrian detector module.
             *\small{\emph{Unable to run on the GPU of the NUC due to a limitation in Tensorflow}}
    }
\end{figure}
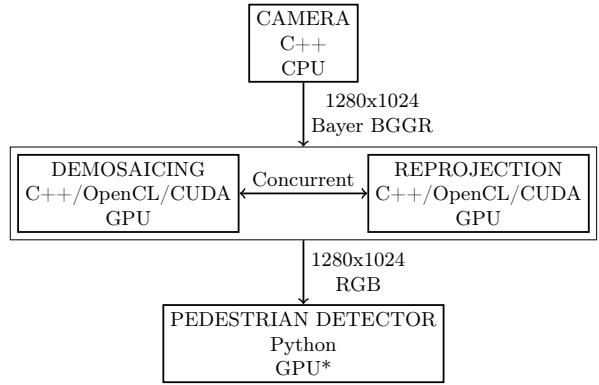

This experiment integrates a DNN into the vision system of a humanoid robotic platform. \cref{fig:pipeline} shows the
vision pipeline that was used in this experiment.

A modular, multi-language software framework, named NUClear~\citep{HoulistonEtAl2016}, is used as the backbone of this
system. Each module in~\cref{fig:pipeline} lists the programming language(s) that were used to implement them.

The camera is fitted with a \ang{195} equidistant field of view lens and streams $1,280\times1,024$ images in a Bayer format
at up to $60$ fps.

The demosaicing and reprojection modules are implemented in OpenCL on the NUC and CUDA on the Jetson and are implemented
such that they are performed concurrently. The output of these modules is a $1,280\times1,024$ RGB image with a
\ang{150} field of view. This means that the reprojection module must also interpolate pixel colour values.

SSD MobileNet~\citep{MobileNet,SSD}, pre-trained on MS COCO~\citep{MSCOCO}, from the Tensorflow Model
Zoo~\citep{TensorflowModelZoo} is used in this experiment.


Due to the use of an equidistant camera lens, the image from the camera has to be projected on to a rectilinear plane
before it can be used as an input to the DNN.\@ However, since the camera lens has a field of view greater than
\SI{\pi}{\radian}, the field of view needs to be restricted, otherwise the resulting rectilinear image would be infinite
in size ($\tan\left(\frac{FOV}{2}\right) = \tan\left(\frac{\pi}{2}\right) = \infty$, see
\cref{eqn:reprojection}).

Reprojecting an equidistant image to a rectilinear image can be performed by first casting a ray from the center of the
camera to a pixel in the rectilinear image and then finding the point that this ray intersects the equidistant image.
In this way, the processing time is determined by the resolution and field of view of the rectilinear image.
\cref{eqn:reprojection} outlines the calculations that are performed for each pixel in the rectilinear image.

\begin{equation}
    \label{eqn:reprojection}
	\begin{split}
        f       &= \frac{\left\lVert{\vec{I}_{wh} - 1}\right\rVert}{2\tan\left(\frac{FOV}{2}\right)} \\
        \vec{v} &= \left\lVert\left< f, \vec{s}_{x}, \vec{s}_{y}\right>\right\rVert
	\end{split}
    \qquad
	\begin{split}
        \theta &= \arccos\left(\vec{v}_{x}\right) \\
        \alpha &= \frac{\theta}{r_{pp}\sin\left(\theta\right)}
	\end{split}
    \qquad
	\begin{split}
        \vec{v}_{out} &=
            \begin{cases}
                \left< 0, 0\right> & \textrm{ if } \theta = 0,\\
                \alpha\vec{v}_{yz} & \textrm{ otherwise}
            \end{cases}
	\end{split}
\end{equation}

Where $\vec{I}_{wh}$ is the width and height of the rectilinear image, $FOV$ is the field of view of the rectilinear
image, $f$ is the focal length of the camera in pixels, $\vec{s}$ is the screen-centered coordinates of a pixel in the
rectilinear image, $r_{pp}$ is a measure of the number of radians spanned by a single pixel in the equidistant image,
and $v_{out}$ is the screen-centered coordinates of a pixel in the equidistant image. The screen-centered coordinate
system places $\left(0, 0\right)$ in the middle of the image, with $+x$ to the right and $+y$ up.

Since Tensorflow does not currently support OpenCL based devices, this experiment is performed with the DNN running on
the CPU of the NUC and the GPU of the Jetson. \cref{fig:pipeline} details which computing device each module runs on.
Moreover, in order to determine the power consumed by the devices, the current drawn from the power supply was recorded
during this experiment.


\section{Results and Observations}
\label{sec:results}
This section details the results of the experiments performed in \cref{sec:experiments}. \cref{sec:res_benchmarks}
details the results of the CPU and GPU device benchmarking. \cref{sec:res_ped} details the results of the
pedestrian detection tasks.

\subsection{CPU and GPU Benchmarking}
\label{sec:res_benchmarks}
\begin{figure}[hp]
	\centering
    \begin{subfigure}{0.49\textwidth}
        \centering
        \includegraphics[width=\linewidth]{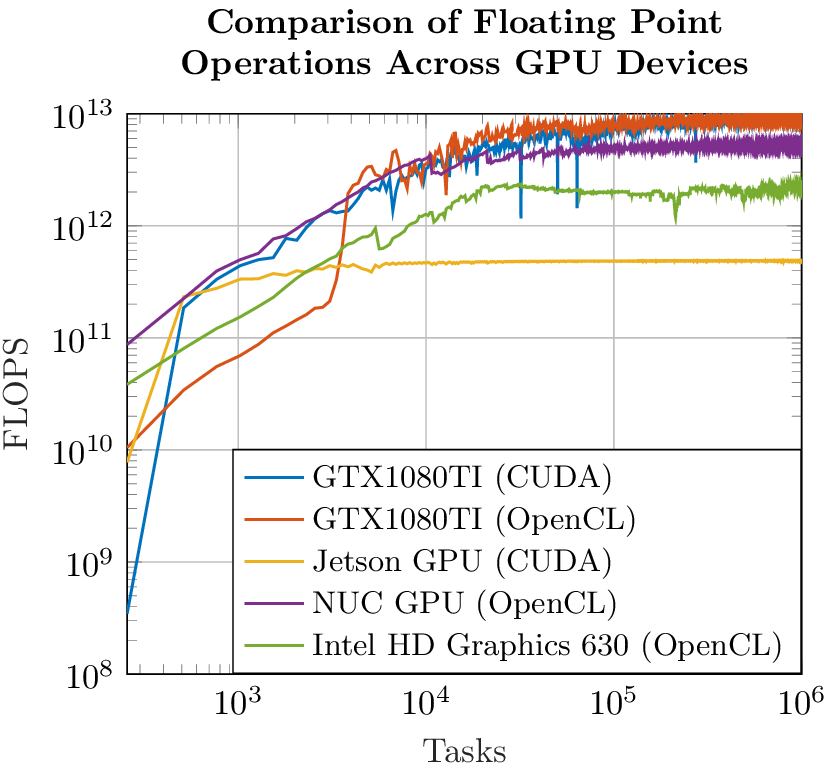}
        \caption{\label{fig:flops_gpu_all}All GPU devices.}
    \end{subfigure}
    \hfill
    \begin{subfigure}{0.49\textwidth}
        \centering
        \includegraphics[width=\linewidth]{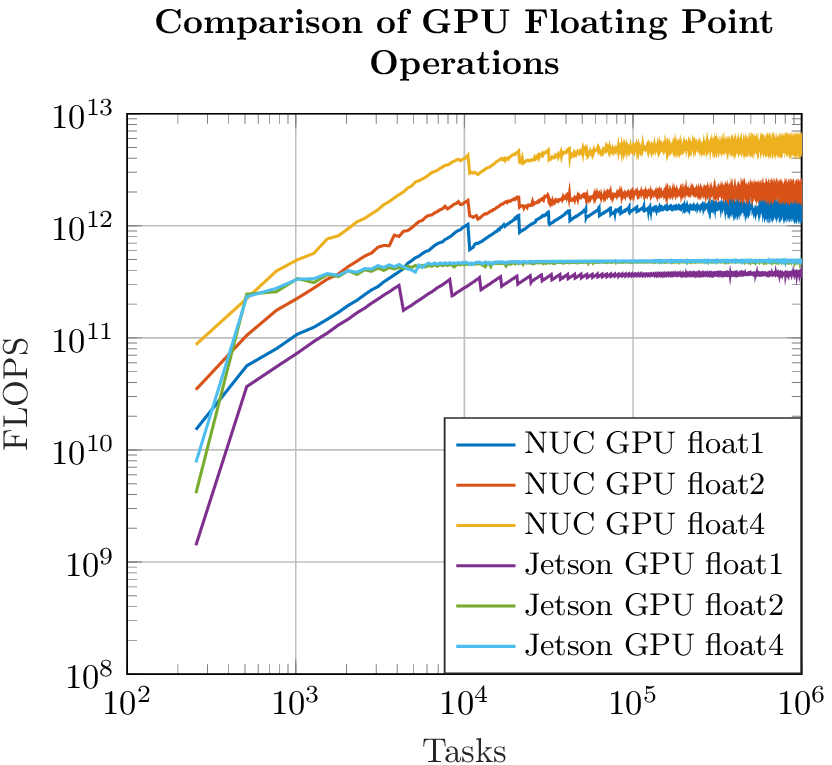}
        \caption{\label{fig:flops_gpu_jetson_vs_nuc}Jetson GPU vs. NUC GPU}
    \end{subfigure}
    \caption{Comparison of FLOPS count for the tested GPU devices.}
\end{figure}

The results of the GPU benchmarking experiment detailed in \cref{sec:exp_benchmarks} provided some surprising results.
\cref{fig:flops_gpu_all} shows the results of implementing and running the benchmarking equations shown in
\cref{sec:exp_benchmarks} using a vector of $4$ single precision floating point values as the main data type. In terms
of performance, we see that the NVIDIA\textregistered~GTX1080Ti performs best, followed by the NUC GPU, the
Intel\textregistered~HD Graphics 630, and finally the Jetson GPU.\@

The interesting part of these results is highlighted in \cref{tab:gpu_devices}, where we see that the calculated
$FLOPS$ value is roughly on par with other reported values for the NVIDIA\textregistered~GTX1080Ti. However, both of the
Intel devices have calculated $FLOPS$ values almost $6$ times higher than other reported values. On the other hand, the
Jetson is calculated to have a $FLOPS$ value roughly $1.5$ times lower than reported.

Unfortunately, it is not currently understood why these values are so wildly out of proportion. It is believed that the
Intel OpenCL drivers have found a way to heavily optimise the kernel, while the reported $FLOPS$ value for the Jetson is
thought to be a theoretical maximum value. This reasoning is backed up by the similar performance of the
NVIDIA\textregistered~GTX1080Ti when running both the OpenCL and CUDA kernels.

\cref{fig:flops_gpu_jetson_vs_nuc} shows a comparison between the NUC GPU and the Jetson GPU, using vectors with
dimensions of $1, 2\text{ and }4$. These results indicate that the Jetson stops providing any performance benefits for
vectors with more than $2$ dimensions. In comparison, the NUC GPU continues to provide performance benefits for vectors
with up to $4$ dimensions. It is also interesting that the $1$ dimension vector performance of the NUC GPU outperforms
the $4$ dimension vector performance of the Jetson.

\begin{table}[htbp]
	\caption{\label{tab:gpu_devices}GPU Devices benchmarked in \cref{sec:exp_benchmarks} and their peak TFLOPS.}
	\centering
	\begin{tabular}{lccc}
		\toprule \\[-0.22cm]
		\textbf{GPU Device}                  & \textbf{OpenCL} & \textbf{CUDA} & \textbf{Reported\footref{ftn:geforce10}\footref{ftn:iris}\footref{ftn:hdgraph}\footref{ftn:jetson}} \\[0.02cm]
		\midrule \\[-0.22cm]
		NVIDIA\textregistered~GTX1080Ti      & $9.14$      & $9.34$      & $10.61$ \\[0.05cm]
		NUC GPU                              & $5.25$      & Unsupported & $0.88$  \\[0.05cm]
		Intel\textregistered~HD Graphics 630 & $2.40$      & Unsupported & $0.44$  \\[0.05cm]
		Jetson GPU                           & Unsupported & $0.49$      & $0.75$  \\[0.05cm]
		\bottomrule
	\end{tabular}
\end{table}

\footnotetext{\label{ftn:geforce10}Wikipedia: GeForce 10 series
	\url{https://en.wikipedia.org/wiki/GeForce_10_series}\addtocounter{footnote}{1}}
\footnotetext{\label{ftn:iris}WikiChip: Intel Iris Plus Graphics 650
	\url{https://en.wikichip.org/wiki/intel/iris_plus_graphics_650}\addtocounter{footnote}{1}}
\footnotetext{\label{ftn:hdgraph}WikiChip: Intel HD Graphics 630
	\url{https://en.wikichip.org/wiki/intel/hd_graphics_630}\addtocounter{footnote}{1}}
\footnotetext{\label{ftn:jetson}Wikipedia: Tegra \url{https://en.wikipedia.org/wiki/Tegra}\addtocounter{footnote}{1}}

\begin{figure}[htbp]
    \centering
    \begin{subfigure}{0.49\textwidth}
        \centering
        \includegraphics[width=\linewidth]{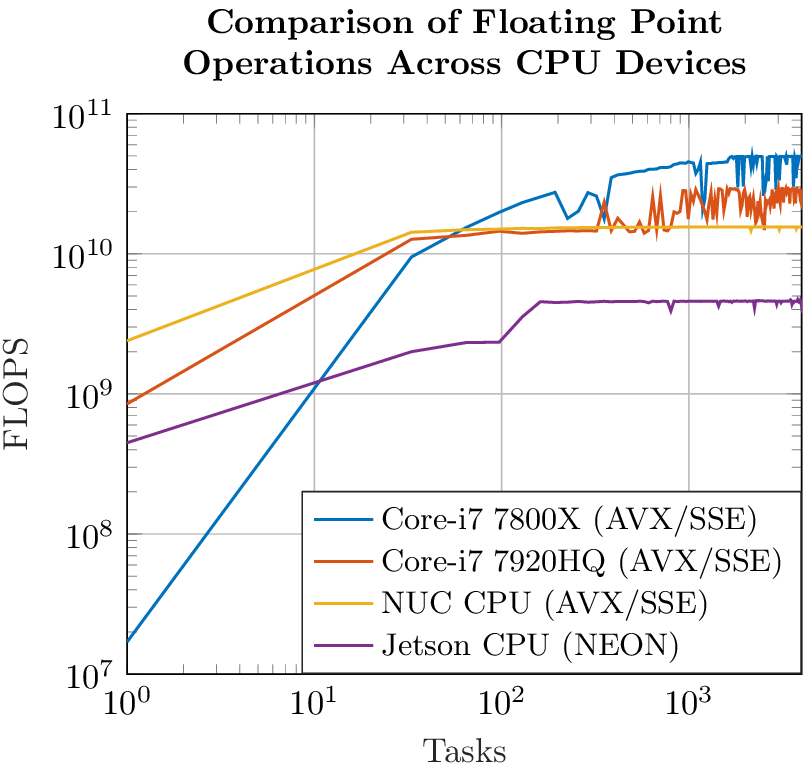}
        \caption{\label{fig:flops_cpu_all}All CPU devices.}
    \end{subfigure}
    \hfill
    \begin{subfigure}{0.49\textwidth}
        \centering
        \includegraphics[width=\linewidth]{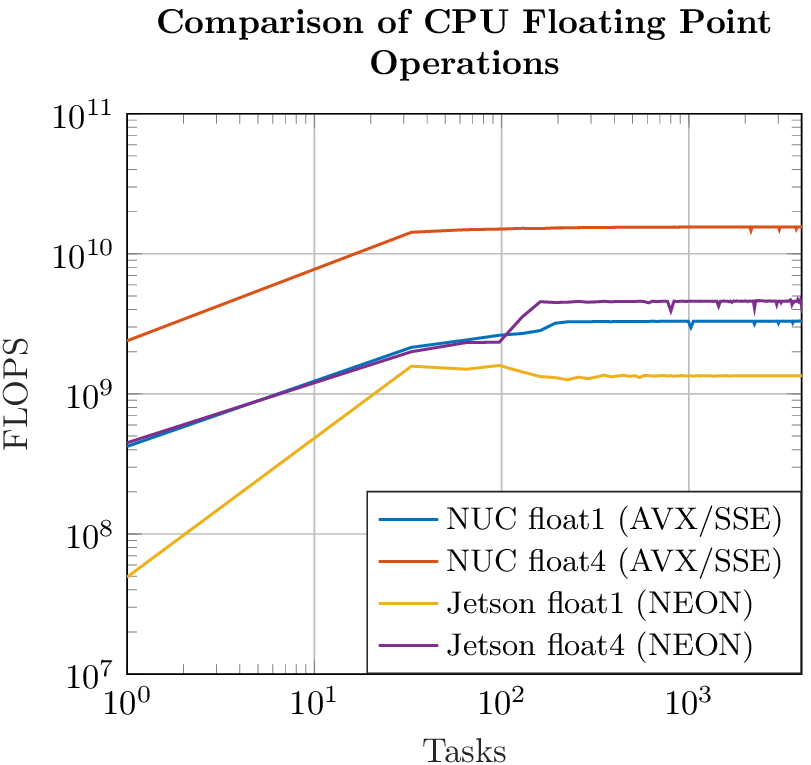}
        \caption{\label{fig:flops_cpu_jetson_vs_nuc}Jetson CPU vs. NUC CPU.
        }
    \end{subfigure}
    \caption{Comparison of the FLOPS count for the tested CPU devices.}
\end{figure}

The results of the CPU benchmarking experiment performed in \cref{sec:exp_benchmarks} also provided some unexpected
results. \cref{fig:flops_cpu_all} shows the performance results of the $4$ CPU devices and \cref{tab:cpu_devices} lists
the calculated FLOPS.\@ Referring to \cref{fig:flops_cpu_all}, the NUC CPU shows the best performance up until 60
tasks are scheduled to run. Above this point, the Intel\textregistered~Core\texttrademark~i7-7800X starts outperforming 
all other devices. The Intel\textregistered~Core\texttrademark~i7-7920HQ shows similar performance to the NUC, and the 
Jetson is approximately $3$ times slower than the NUC CPU.\@ The erratic performance of the
Intel\textregistered~Core\texttrademark~i7-7800X and 7920HQ are due to these desktop/laptop devices running other 
tasks during the run time of the test.

\begin{table*}[htbp]
	\caption{\label{tab:cpu_devices}CPU devices benchmarked in \cref{sec:exp_benchmarks} and their peak FLOPS.}
	\centering
	\begin{tabular}{llcc}
		\toprule \\ [-0.22cm]
		\textbf{CPU Device}                               & \textbf{CPU Source}              & \textbf{Intrinsics} & \textbf{GFLOPS} \\ [0.05cm]
		\midrule \\ [-0.22cm]
		Intel\textregistered~Core\texttrademark~i7-7800X  & Dell Alienware Area-51           & AVX/SSE             & $49.68$         \\ [0.05cm] 
		Intel\textregistered~Core\texttrademark~i7-7920HQ & Apple MacBook Pro 14,3           & AVX/SSE             & $29.33$         \\ [0.05cm] 
		NUC CPU                                           & Intel\textregistered~NUC7i7BNH   & AVX/SSE             & $15.54$         \\ [0.05cm]
		Jetson CPU                                        & NVIDIA\textregistered~Jetson TX2 & NEON                & $4.69$          \\ [0.05cm]
		\bottomrule
	\end{tabular}
\end{table*}

\subsection{Image Reprojection, Demosaicing, and Pedestrian Detection}
\label{sec:res_ped}
The results of the experiment run in \cref{sec:exp_ped} show that for the purposes of reprojecting and demosaicing a
high-resolution image, the GPU in the NUC and the Jetson are closely matched, with the NUC completing the preprocessing
tasks in \SI{6.21}{\milli\second} ($SD = 0.99$) compared to the Jetson's time of \SI{6.48}{\milli\second} ($SD = 2.43$).
However, with respect to running a deep neural network for pedestrian detection on high resolution images, the CPU in
the NUC is significantly faster than the GPU of the Jetson (by a factor of $3.5$), with the NUC completing the task in
\SI{0.17}{\second} ($SD = 0.19$) and the Jetson completing the task in \SI{0.57}{\second} ($SD = 1.10$).

It should be noted that the performance times for the Jetson include the time needed for transferring the image to the
GPU and then transferring the results back from the GPU.\@ Such transfer times are not part of the NUC performance times
as the pedestrian detection was carried out on the NUC's CPU.\@ Based on the performance times for both the NUC and
Jetson on the image demosaicing and reprojection tasks, it would not be unreasonable to conclude that the data transfer
times involved in these tasks are not significant enough to drastically alter the results that are seen here.

\begin{figure}[htbp]
    \centering
    \begin{subfigure}{0.32\textwidth}
        \centering
        \includegraphics[width=\linewidth]{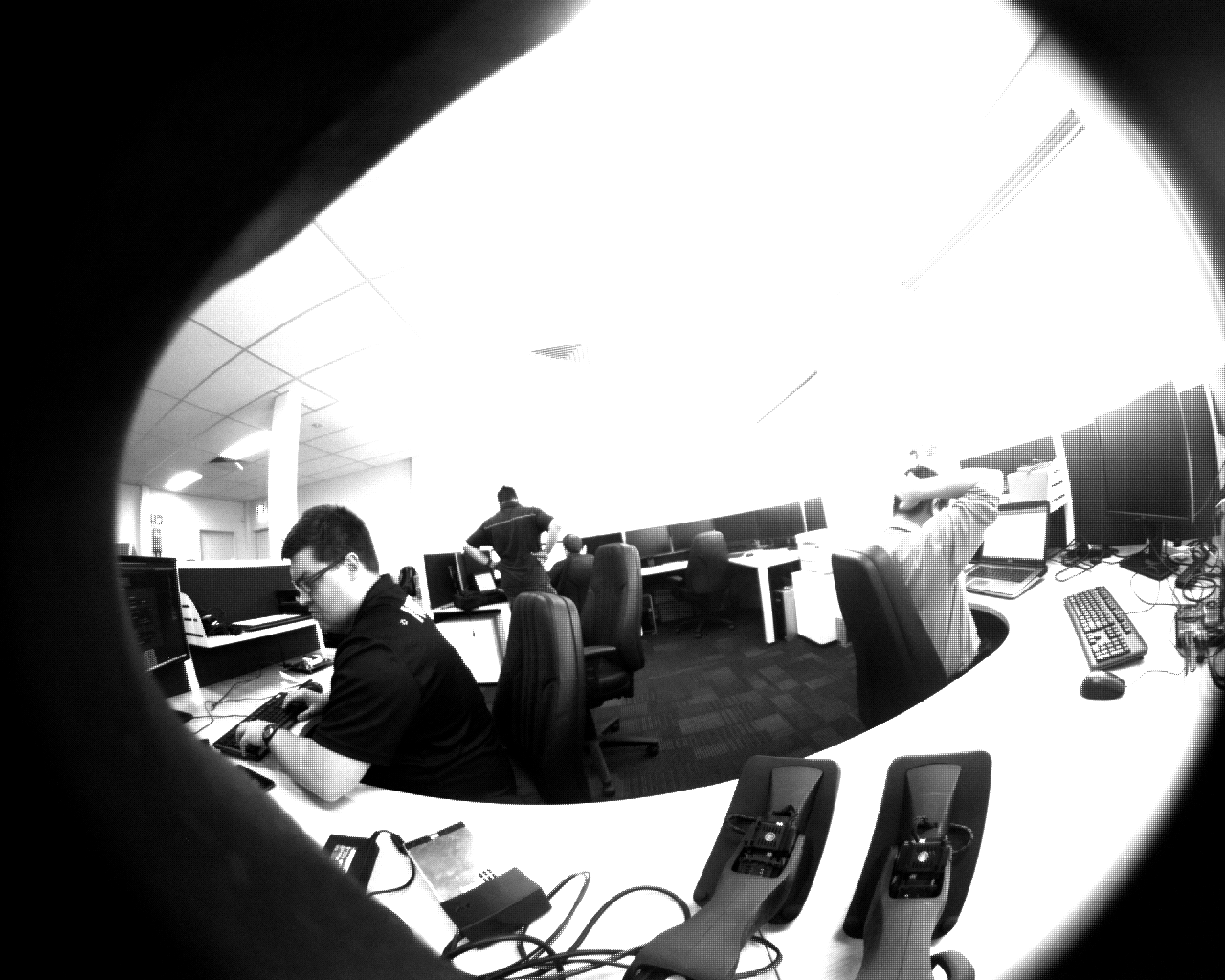}
    \end{subfigure}
    \hfill
    \begin{subfigure}{0.32\textwidth}
        \centering
        \includegraphics[width=\linewidth]{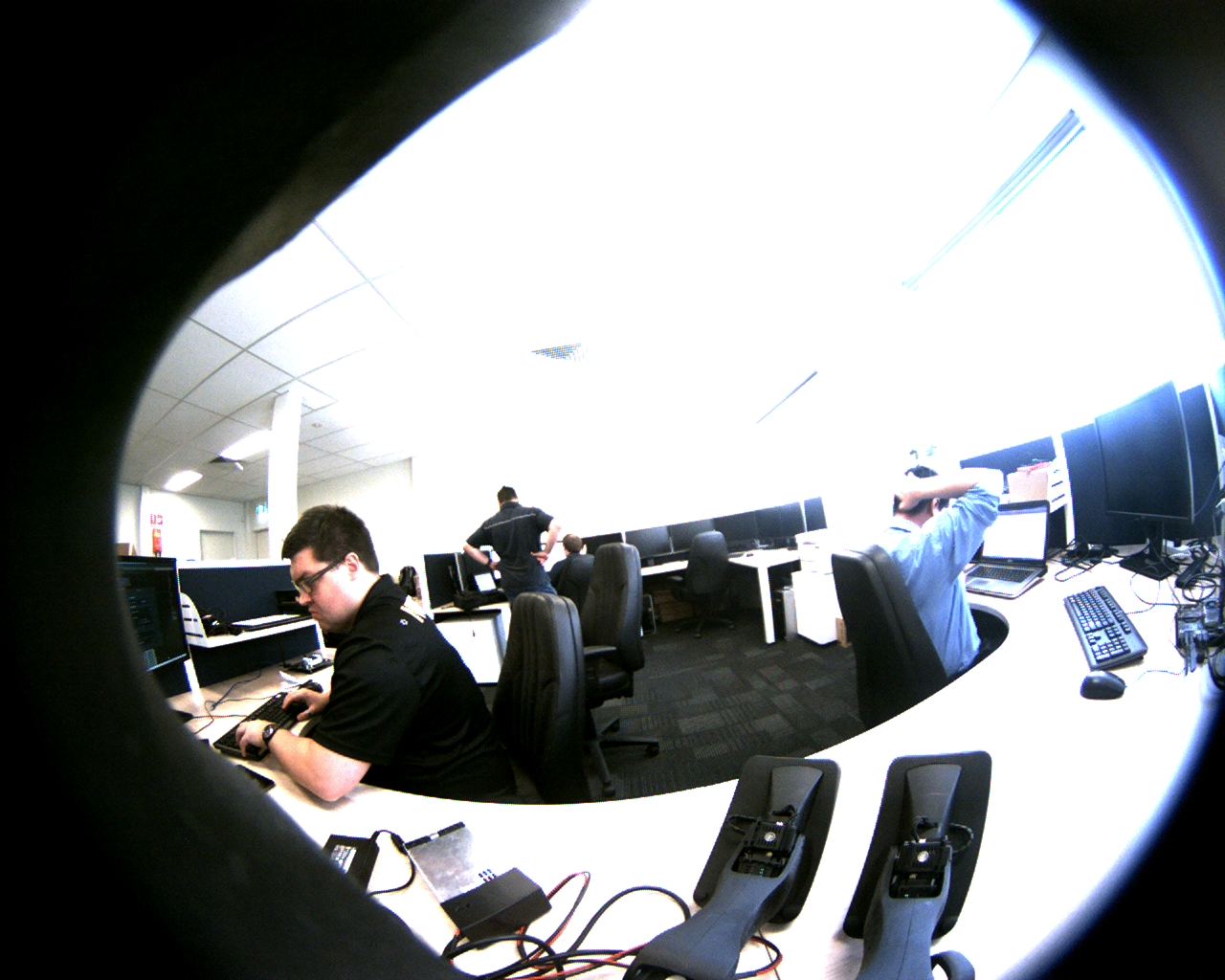}
    \end{subfigure}
    \hfill
    \begin{subfigure}{0.32\textwidth}
        \centering
        \includegraphics[width=\linewidth]{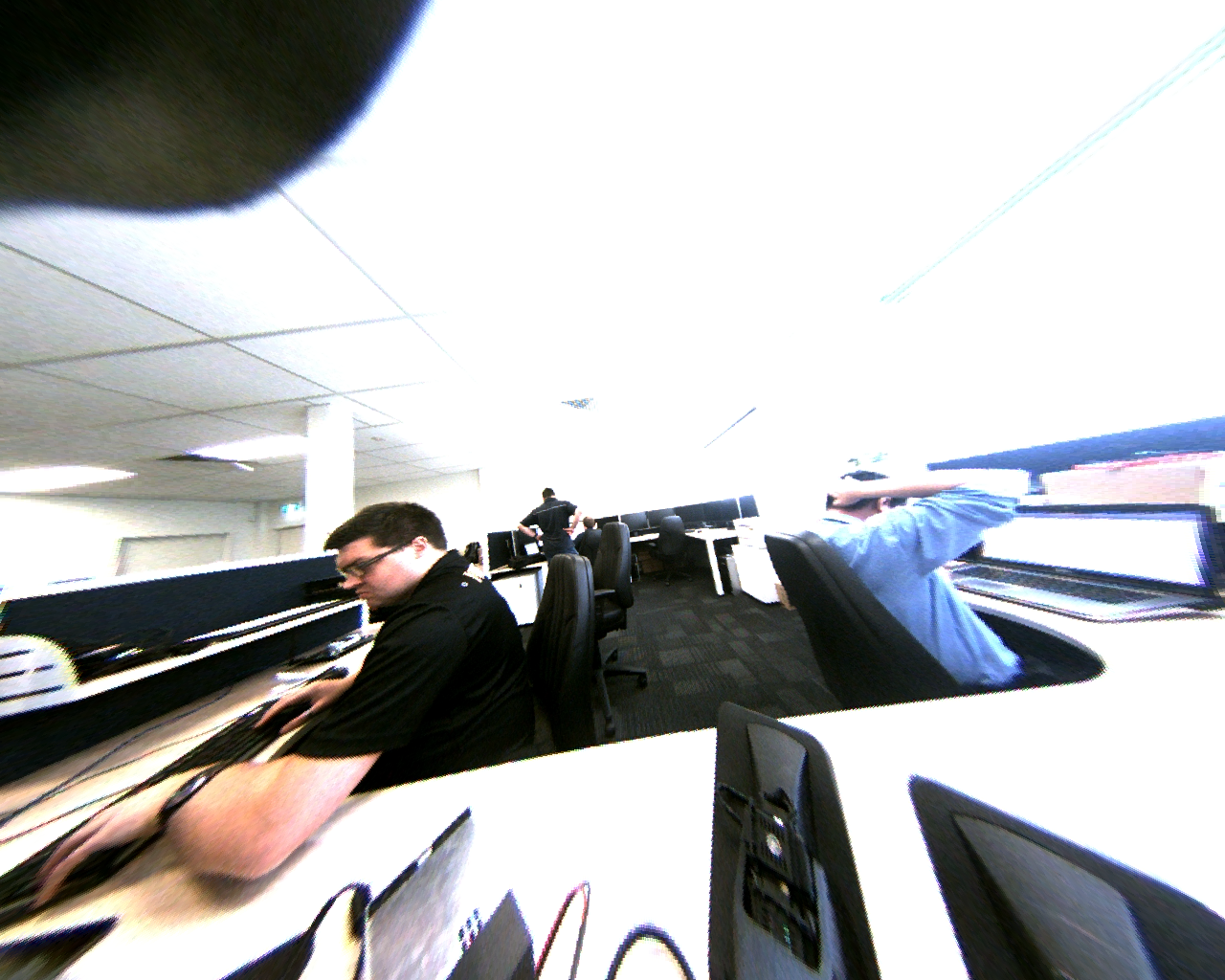}
    \end{subfigure}
    \caption{\label{fig:res_reproject}
        Results of the image reprojection algorithm.
		Left: Original mosaiced image, interpreted as greyscale, with \ang{195} FOV.\@
		Middle: Demosaiced image.
		Right: Demosaiced and reprojected image with \ang{150} field of view.
        The black border on the left side of the two images on the left is due to the robot's nose.
        All images have a resolution of $1,280\times1,024$ pixels.
    }
\end{figure}

\cref{fig:res_reproject} shows the results of demosaicing and reprojecting an image from the camera. The pinched black
shape on the left side of the first two images in \cref{fig:res_reproject} are due to the robot's nose obstructing the
field of view. These two images have a \ang{195} equidistant field of view, while the last image has a \ang{150}
rectilinear field of view. All images have a resolution of $1,280\times1,024$ pixels (not shown to scale).

\begin{figure}[htbp]
	\centering
	\includegraphics[width=0.49\textwidth]{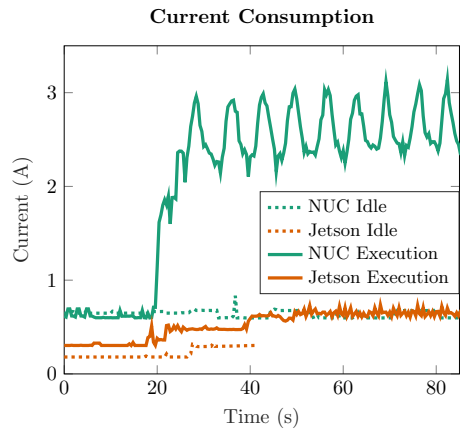}
	\caption{\label{fig:current}Comparison of current consumption for the Jetson and NUC.}
\end{figure}

\cref{fig:current} shows the current consumption on the NUC and the Jetson. On average, we see that the NUC draws
\SI{2.53}{\ampere} ($SD = 0.35$), while the Jetson is drawing \SI{0.59}{\ampere} ($SD = 0.09$). When powered by a
\SI{16}{\volt} power supply we see that the NUC has a power consumption of \SI{40.52}{\watt}, compared to the Jetson's
\SI{9.48}{\watt}. Considering the maximum $FLOPS$ determined in \cref{sec:res_benchmarks} and the maximum current draw,
we see that although the NUC has a larger power requirement, we achieve a $2.5$-fold increase in the number of $FLOPS$
available per watt of power consumed, with the NUC providing $104.64$ GFLOPS/W compared to the Jetson's $41.26$ GFLOPS/W.

If we were to power these devices from a \SI{14.8}{\volt} $4$-cell LiPo battery with a \SI{3850}{\milli\ampere\per\hour}
capacity we could expect a battery life of \SI{84.36}{\minute} on the NUC and \SI{360.48}{\minute} on the Jetson. These
figures are calculated based on the average current draw for the two devices, while also assuming that as system voltage
decreases current draw will increase in order to keep system power consumption constant.

\section{Discussion and Conclusion}
This work presented a series of performance tests of computer vision tasks on two computer platforms -- the NUC and 
the Jetson. Despite having a larger power requirement, the NUC proves to be a powerful computing device for an embedded
platform. It should be noted that, if Tensorflow was compiled from source for the NUC, further performance improvements
are expected as the NUC supports SIMD intrinsics which the pre-built Tensorflow is not compiled to use. Similar
performance improvements would not be expected on the Jetson, as Tensorflow is using the GPU device on this platform
with very little work being done on the CPU.\@ More significant improvements would be expected on the NUC if Tensorflow
was compiled to use OpenCL as a backend, as opposed to CUDA.\@ This would allow Tensorflow to fully utilise the GPU on
the NUC.\@ If the intended system has a severely constrained power budget ($<$ \SI{10}{\watt}), then the Jetson becomes
a very strong candidate for embedded GPU computing tasks.

We conclude that for utilising deep networks on a humanoid robot, like the NUgus, a careful selection of the computing
platform is essential. The initial assumption that the GPU focused Jetson would be the best platform for our robot was
not supported by this comparative evaluation. It will be interesting to see what opportunities for progress in machine
learning on robots the next generation of computing devices may bring.

%
%
\bibliographystyle{splncs04}
\bibliography{robotics}
\end{document}